\pdfoutput=1

\documentclass[11pt]{article}

\usepackage[final]{acl}

\usepackage{times}
\usepackage{latexsym}
\usepackage{graphicx}
\usepackage{amsmath}

\usepackage[T1]{fontenc}

\usepackage[utf8]{inputenc}

\usepackage{microtype}

\usepackage{inconsolata}

\definecolor{km}{HTML}{FF0000}

\title{\textsc{StorySumm}: Evaluating Faithfulness in Story Summarization}

\author{Melanie Subbiah$^*$\\
  Columbia University \\
  \texttt{m.subbiah@columbia.edu} \\\And
  Faisal Ladhak$^*$ \\
  Answer.AI \\
  \texttt{fl@answer.ai}
  \\\And
  Akankshya Mishra \\
  Columbia University \\
  \texttt{am6203@columbia.edu}
    \\\AND
  Griffin Adams \\
Answer.AI \\
  \texttt{ga@answer.ai}
    \\\And
  Lydia B. Chilton \\
  Columbia University \\
  \texttt{chilton@cs.columbia.edu}
    \\\And
  Kathleen McKeown \\
  Columbia University \\
  \texttt{kathy@cs.columbia.edu} \\}

\begin{document}
\maketitle
\def\thefootnote{*}\footnotetext{These authors contributed equally to this work.}\def\thefootnote{\arabic{footnote}}
\begin{abstract}
Human evaluation has been the gold standard for checking faithfulness in abstractive summarization. 
However, with a challenging source domain like narrative, multiple annotators can agree a summary is faithful,
while missing details that are obvious errors only once pointed out. We therefore introduce a new dataset, \textsc{StorySumm}, 
comprising LLM summaries of short stories with localized faithfulness labels and error explanations. This benchmark is for evaluation methods, testing whether a given method can detect challenging inconsistencies. Using this dataset, we first show that any one human annotation protocol is likely to miss inconsistencies, and we advocate for pursuing a range of methods when establishing ground truth 
for a summarization dataset. We finally test recent automatic metrics and find that none of them achieve more than 70\% balanced accuracy on this task, demonstrating that it is a challenging benchmark 
for future work in faithfulness evaluation. 
\end{abstract}

\section{Introduction}

As Large Language Models (LLMs) are able to perform more open generation tasks
, challenges in evaluation have arisen \cite{gabriel2020go}. Summarization is one such task.
 Some aspects of summary quality like readability or coherence \cite{goyal2022snac,chang2023booookscore} can be judged by looking at the summary alone. However, judging faithfulness (whether all details in the summary are faithful to the source) requires carefully checking a multi-sentence summary against a multi-paragraph source document \cite{krishna2023longeval}. Summaries that misrepresent source documents can easily spread disinformation, so it is critical we evaluate summary faithfulness, despite how labor-intensive it is.

\label{sec:intrometrics}Methods for detecting inconsistencies have generally used one of two tools: 1) trained models, or 2) human crowdworkers. 
Model-based approaches typically build on QA or entailment strategies. QA strategies generate questions about the summary and compare answers retrieved from the summary vs. the source document \cite{durmus2020feqa, fabbri2021qafacteval}. Entailment-based approaches align facts in the summary with evidence from the source and determine for each pair if the evidence entails the fact \cite{utama2022falsesum, laban-etal-2022-summac, maynez2020faithfulness}. More recent work explores prompting strategies for LLMs to identify faithfulness errors \cite{min2023factscore, kim2024can, si2023large, luo2023chatgpt, manakul2023selfcheckgpt}. 

\begin{figure}[t]
    \centering
    \includegraphics[width=\columnwidth]{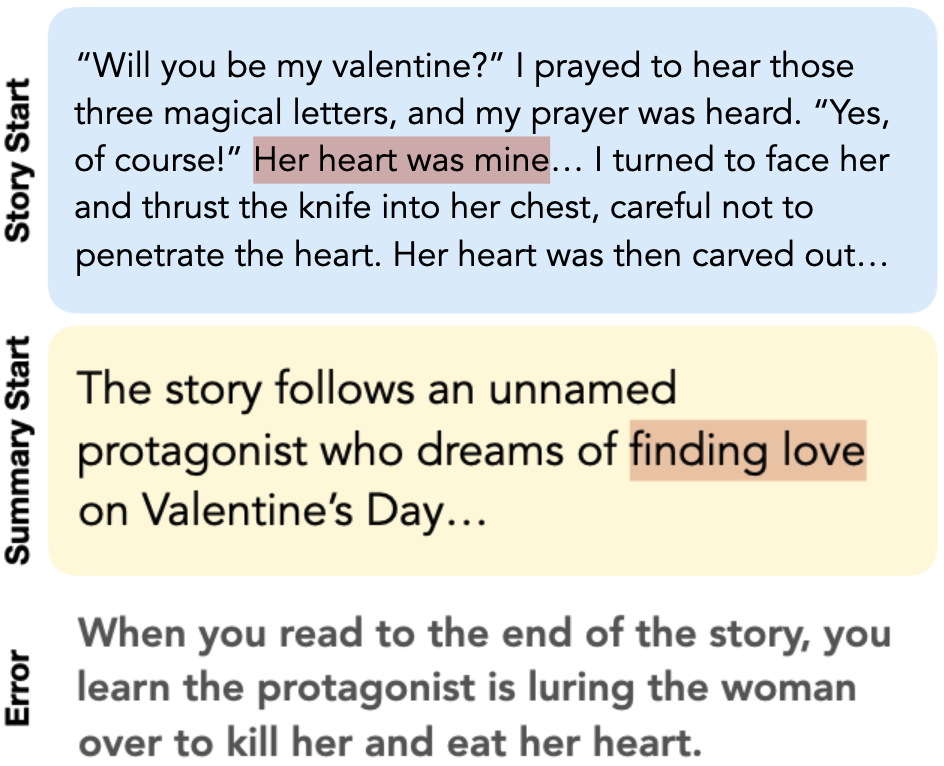}
    \caption{A \textsc{StorySumm} example illustrating an incorrect interpretation of double entendre. A standard fine-grained human annotation protocol missed this inconsistency even though it is obvious once pointed out.}
    \label{fig:heart}
\end{figure}

With human annotators, prior work has shown that human judgments have increased variability when evaluating long summaries \cite{krishna2023longeval}. Reducing the problem to evaluating individual sentences or claims helps to produce more reliable results \cite{krishna2023longeval, ye2023flask, min2023factscore}. However, these works have focused on factuality in news summaries or real-world articles where ground truth is based in reality and facts are stated explicitly.

As LLMs continue to grow in capabilities, there is a pressing need for evaluation of their accuracy to grow with them. 
We therefore produce a new benchmark, \textsc{StorySumm}, which can be used to improve evaluation methods for faithfulness. \textbf{\textsc{StorySumm} consists of 96 short stories and LLM-generated summaries with over 500 sentence-level faithfulness labels and explanations}. Each unfaithful summary is labeled as \textit{easy} or \textit{hard} to detect. 

LLM summaries
often contain subtle errors, particularly for narrative text which requires nuanced interpretation. This benchmark therefore introduces new challenges
when compared to fact-checking or summarization datasets in the news domain. 
 The example in Figure \ref{fig:heart} demonstrates that assessing the summary requires correct interpretation of sentences like, \textit{Her heart was mine.}, which have multiple meanings and are misleading without carefully reading the entire story
 . By focusing on faithfulness in narrative summarization and using real-world data from LLMs and Reddit, 
\textsc{StorySumm} poses a realistic but hard benchmark to push our methods forward.


We first explore how to establish ground-truth on this dataset by comparing different human annotation protocols and manually inspecting the results. We try different protocols and pools of annotators to see if there is an approach that helps average annotators pay attention and understand this challenging task more consistently. We find that different protocols catch unique but legitimate inconsistencies and have only fair agreement with each other. We therefore manually review and merge label sets across three annotation protocols.

We analyze the errors found by each protocol, and formulate a set of \textbf{recommendations for human evaluation} of faithfulness in narrative summarization. Most importantly, we show that \textbf{it is important to use a variety of annotators and protocols when establishing ground truth for faithfulness}. We then explore how well recent automatic metrics perform on this dataset. We find that \textbf{no metric achieves more than 70\% balanced accuracy} on this task and even the best metric misses almost 50\% of the hard inconsistencies.\footnote{All code and data will be released at \href{https://github.com/melaniesubbiah/storysumm}{https://github.com/melaniesubbiah/storysumm}.}

\section{\textsc{StorySumm} Dataset}

We design our benchmark with a focus on three principles which distinguish it from existing datasets. First, the stories need to be short enough that humans can easily read them, so that we can affordably test human protocols. 
Second, the stories should not be so famous that LLMs have likely trained on summaries of them, potentially biasing LLM summary or evaluation quality. Third, the summaries should be representative of powerful LLMs so that we can assess how difficult it is to find errors in fluent and convincing summaries. 

\begin{table}[t]
\centering \small
\begin{tabular}{l|c|c|c} 
Split & \# & Sto. wc & Sum. wc \\\hline\hline
Val. & 33 & 610 & 120\\
Test & 63 & 849 & 149\\
All & 96 & 767 & 139 \\
\end{tabular}
\caption{Summary statistics for \textsc{StorySumm} showing the number of story-summary pairs and the average word count of stories and summaries.}
\label{tab:datastats}
\end{table}

Motivated by these principles, we opt for short narratives from Reddit and use GPT-series and Claude-series models to generate summaries. We do not include any human-written summaries as the purpose of this dataset is to improve detection of errors in LLM-generated summaries. 
 We show summary statistics for the dataset in Table \ref{tab:datastats} and full examples of stories/summaries in Appendix \ref{sec:appendixexamples}.

\subsection{Stories}

We collect a dataset of 32 short stories from two popular subreddits where users can submit their original short stories for others to enjoy and comment on.\footnote{The two subreddits we used were \href{https://www.reddit.com/r/shortstories/}{r/shortstories} and \href{https://www.reddit.com/r/shortscarystories/}{r/shortscarystories}.} We filter out posts that are marked NSFW (Not Safe For Work, meaning inappropriate content) and also posts that have fewer than three up-votes. The stories are typically less than one page long. We note that users do not write summaries for their stories, and since these stories are not popular, they're unlikely to be summarized elsewhere; therefore, there is little concern about data contamination for LLMs. Additionally, as LLMs are now being used to summarize lots of different data online, it is important to evaluate them on more colloquial narrative like this rather than just benchmarks of published/popular stories. 

\subsection{Summaries}

For each story, we generate 3 different summaries using 3 different models, resulting in 96 story-summary pairs (see Appendix \ref{sec:prompts} for prompting details)\footnote{Accessed through the \href{https://platform.openai.com/playground}{OpenAI API} and \href{https://console.anthropic.com/}{Anthropic API}.}. Each summary is about a paragraph long. To simulate real evaluation conditions, we split the dataset into a validation split of 33 summaries which are generated by an older set of models (Davinci-3, ChatGPT, and Claude-2) and a test set of 66 summaries from newer models (GPT-3.5, GPT-4, Claude-3). This allows us to assess whether automatic metrics that require threshold tuning for classification can be tuned on a validation set of labeled summaries from older models and still work well as newer models are coming out. We use disjoint sets of 11 and 21 stories to generate the summaries for the validation and test sets respectively.


 \subsection{Annotator Labels}

The question we ask annotators is: \textit{Is the information in the summary consistent with the story?} We define a \textit{consistent} summary as: \textit{The events and details described in the summary should not misrepresent details from the story or include details that are unsupported by the story. We ask you to ignore commentary in evaluating consistency.} 
Commentary means sentences like, \textit{The story reflects the enduring bonds of friendship and the role of companionship during times of hardship.}, which interpret the story to find themes rather than just detail the plot.

For annotator recruitment\footnote{Our work with annotators is approved by Columbia University IRB protocol AAAS4051
.}, we first compare Amazon Mechanical Turk and Upwork\footnote{\href{https://www.mturk.com}{MTurk} and \href{https://www.upwork.com}{Upwork}}, asking four annotators from each platform to assign a binary faithful/unfaithful label to each summary. We mark a summary as faithful if three or more annotators in a group label it as such. We find that MTurk workers label 97\% of summaries faithful whereas Upwork workers label 64\% as faithful. When the authors perform the same task, we find 45\% of summaries faithful, so we conclude that Upwork workers are more astute at catching errors and we use them for the remainder of our experiments. We caution future work to avoid using MTurk for faithfulness evaluation as it will dramatically inflate performance. \citet{marshall2023broke} also showed Mturk response quality has dramatically declined in the last decade and is now mostly unusable.

We build on \citet{krishna2023longeval}, which shows that fine-grained evaluation encourages inter-annotator agreement. We recruit three annotators from Upwork who are fluent in English and successfully complete a pilot exercise shown in Appendix \ref{sec:pilot}. We then ask them to assign a binary faithfulness label to each sentence in a summary. When they mark a sentence as unfaithful,
they also provide a brief written justification. 
Prior work has shown limited benefit to using atomic claims \cite{tang2024minicheck}, so we do not take the additional step of generating them as proposed by \citet{min2023factscore}. The full interface for experiments is shown in Appendix \ref{sec:interface}. We pay each annotator \$100 for annotation of all 96 summaries.

If two or more annotators mark the same sentence as unfaithful, we mark the whole summary as unfaithful. If all three annotators mark the same sentence as unfaithful, we label that unfaithful summary as \textit{easy} to detect, whereas it is \textit{hard} to detect if one annotator labels the sentence as faithful. These difficulty distinctions allow for more meaningful error analysis of different evaluation methods, and only apply to 
unfaithful summaries. We see almost perfect inter-annotator agreement with a Fleiss-kappa score of 0.85 for the sentence-level annotations. 

   \begin{table}[t]
\centering \small
\begin{tabular}{l|c|c|cc}
Generator & \# & \% Faith. & \# Easy & \# Hard\\\hline\hline
Davinci-3 & 11 & 72.7\% & 1 & 2\\
GPT-3.5 & 21 & 57.1\% & 3 & 6 \\\hline
ChatGPT & 11 & 54.5\% & 4 & 1\\
GPT-4 & 21 & 57.1\% & 2 & 7\\\hline
Claude-2 & 11 & 36.4\% & 4 & 3\\
Claude-3 & 21 & 90.5\% & 0 & 2\\\hline
\textbf{Overall} & 96 & 63.5\% & 14 & 21\\
\end{tabular}
\caption{Summary statistics using the annotator labels for each summary generation model, showing the number of summaries, the percent of summaries labeled faithful, and the number of unfaithful summaries labeled \textit{easy}/\textit{hard} to detect.}
\label{tab:modelscores}
\end{table}
 
 Finally, by the breakdown of faithfulness labels by model shown in Table \ref{tab:modelscores}, we can see that faithfulness is still a significant problem for LLMs in narrative summarization with close to 40\% of summaries containing errors. For example, Table \ref{tab:example_error_all} shows a case where all three models misinterpret the protagonist as having a positive reaction to speaking to their dead mother when in reality the mother was very cruel in life. \citet{kim2024fables} and \citet{subbiah2024reading} also find many errors in interpretation of characters' internal states in narrative summaries like this example. 
 
 Table \ref{tab:modelscores} also shows a general trend of newer models having more \textit{hard} unfaithful summaries and older models having more \textit{easy} ones. This pattern suggests that model errors will continue to become harder to detect in subsequent generations. We show more \textsc{StorySumm} examples in Table \ref{tab:easyhardexamples}. 

 \begin{table}[t]
\centering \small
\begin{tabular}{|p{7cm}|}
\hline\textbf{Story Evidence: }...I remember Mom, a sweet woman. Her smile, her strong hands, and witty humor . . .
After several minutes, I gather the courage to speak to her.
“How’s it going, mom?” I feel strong emotions coming.
Mom’s voice answers me in return:
You brat! I took care of you and your brother! I gave you your own closet. But you scratched it like a cat! You! You didn’t know how to wash in the tub, so I showed you how!
I laugh at the whisper. “You’re telling a different story, mom.” I smile. “At least you won’t hurt me or Hector anymore.”\\\hline\\
\textbf{Davinci-3: }...The narrator talks to their mother's grave and  \color{red}{reminisces, and finds that even in death, their mother still has plenty of wit and humor.}\\\\
\textbf{ChatGPT: }...They eventually make it to their mother's grave and have a conversation with her, \color{red}{reminiscing about their past. The story ends with the protagonist feeling comforted by their mother's voice.}\\\\
\textbf{Claude-2: }...The narrator speaks to the mother's spirit and  \textcolor{red}{reminisces about memories of her. Despite the strange ability, the narrator finds comfort communicating with the mother's spirit}...\\\hline
\end{tabular}
\caption{An example where all three LLMs make the same error (shown in red) in understanding the narrative. The protagonist had a cruel relationship with their mother, but the summaries all suggest the protagonist feels positively about speaking with her ghost.} 
\label{tab:example_error_all}
\end{table}

 \begin{table*}[t]
\centering \small
\begin{tabular}{l|p{4.5cm}p{5.5cm}p{3.35cm}} 
& \textbf{Story Evidence} & \textbf{Summary Claim} & \textbf{Reason for Error}\\\hline\hline
 \textbf{Easy} & \scriptsize He woke up staring at a bright florescent light. He could hear his father talking to the doctors and police. Daniel thought it was best to stay quiet. & When he wakes up he is in a hospital and his parents are discussing sending him to rehab. \textcolor{red}{Daniel agrees}, and then falls back asleep. & \textit{Daniel does not agree to go to rehab the first time he wakes up in the hospital}.\\\cline{2-4}
& \scriptsize I could still taste the gas station coke I had slurped up before the light pulled me into the night sky. In what felt like seconds, I was swallowed up in a beam of light. &  A man is abducted \textcolor{red}{from his car} while drinking a soda by a beam of light. & T\textit{here is no evidence the man was in his car.}\\\cline{2-4}
& \scriptsize Kristen's Dad and her little brother Christian sat quietly... "Dad what is going to happen to Kristen?" Christian asked. Her Dad did not respond, and continued to slowly eat. & Her father and \textcolor{red}{younger brother know what's happening}, but they are unable to stop it... & \textit{Her younger brother does not know what is happening and is asking their dad.}\\\cline{2-4}
& \scriptsize Aiming under my own chin, I pulled the trigger. I didn't hear the blessed scream of the barrel. & \textcolor{red}{They contemplate ending their life}, but instead their memory is wiped... & \textit{The narrator takes action to end their life.}\\\cline{2-4}
& \scriptsize But she, along with her strange tubes and tanks and half-smiles was gone. The last thing he remembered seeing yesterday, while he was halfway across the street, was a blaring alarm and a screeching van, (red? white?). & But one day, the girl doesn't show up and \textcolor{red}{he learns that she has been taken to the hospital}. & \textit{The "he" is a dog and so doesn't know the girl is at the hospital but the reader can infer it.}\\\cline{2-4}
& \scriptsize Eventually on the road I met a couple travelers who were all too happy to trade me 3 silver coins for my gold coin. & Eventually, after trading one of his gold coins \textcolor{red}{for 15 silver ones}, he wakes to find his previously tiny dragon grown bigger... & \textit{The narrator makes a trade for only 3 silver coins.}\\\hline\hline
\textbf{Hard} & \scriptsize Margot starts gathering the plastic white discs. One by one, I frantically pitch the AirTags out the open window into the speeding gravel, each shattering on impact. & Jane \textcolor{red}{forces Margot to throw} the AirTags out the window, concerned for her safety. & \textit{Jane (the narrator) throws the AirTags out the window.}\\\cline{2-4}
Annotator & \scriptsize It is said that men have trouble listening to women.  I had no trouble listening to my mom. & The author... recalled \textcolor{red}{her} mother once speaking about how she used to love eating honeycombs... When \textcolor{red}{he} hints at the gift... & \textit{The narrator is a man, but the summary uses a mix of pronouns for the narrator.}\\\hline
& \scriptsize "There's only one plate," she said, puzzled... I turned to face her and thrust the knife into her chest, careful not to penetrate her heart... Her heart was carefully set on the plate... & He prepares a meal \textcolor{red}{for her}, but when she arrives, he stabs her and carves out her heart to eat. & \textit{The meal is only for himself to eat her heart.}\\\cline{2-4}
Expert & \scriptsize "I'll see you tomorrow." The way she said tomorrow-... Naturally, he assumed she would say nothing else but his name with such emotion. The small terrier knew, like the sky is blue, that his name was Tomorrow. & In this poignant story, a small terrier \textcolor{red}{named Tomorrow} has been living on the streets for as long as he can remember. & \textit{The reader infers that the dog is not actually named Tomorrow.}\\\hline
& \scriptsize The guy looked taken aback, "Ma'am, I have a husband who I am completely devoted to!..." The guy's husband looked back at her, "Get lost, Karen..." & She then descends to the mortal realm to test her power, but is \textcolor{red}{turned down by a gay man who calls her a Karen}. & \textit{The man's husband calls her Karen.}\\\cline{2-4}
Hybrid & \scriptsize He said he never got the opportunity to use it, and apologized again. The rescue team looked at each other, just as the radio flared to life... & When the team was \textcolor{red}{about to leave}, the radio came to life again and the same voice asked when they were coming to get him. & \textit{There is no evidence the team was about to leave.}\\\hline
\end{tabular}
\caption{Examples from \textsc{StorySumm}. The easy examples are detected by all three annotators in the annotator labels and all three human annotation methods (annotator, expert, and hybrid). The hard examples are detected only by the method listed on the left. We present evidence from the story, the erroneous summary claim, and the error reason.}
\label{tab:easyhardexamples}
\end{table*}

\section{All That Glitters is Not Gold...}


Typically, annotator labels with almost perfect inter-annotator agreements like ours are just assumed to be ground truth. However, we hypothesize that errors in narrative summaries may be difficult to catch and the annotators likely missed some. Therefore, we compare our Upwork annotator labels against labels from other human evaluation protocols to gain a better sense of their quality. In addition to our annotator labels, we compare the following two methods\footnote{See Appendix \ref{sec:moreresults} for additional methods we experimented with but rejected.}:
 
\paragraph*{Expert.} Three of the authors of this paper review each summary and label it as faithful or unfaithful. We consider ourselves ``expert" annotators as we have experience in faithfulness research and are motivated to produce thoughtful labels (modeled after \citet{kryscinski2019evaluating} who also use themselves as expert annotators for factual consistency).
The three experts adjudicate their labels by discussing any disagreements until all three agree on the label. This process is completed before the experts view any other labels for the dataset so they can remain unbiased. 
All three experts initially agreed on only 46\% of the labels before adjudication, demonstrating that even experts struggle to catch every error. In total, this process took about ten hours.

\begin{figure}[t]
    \centering
    \includegraphics[width=.9\columnwidth]{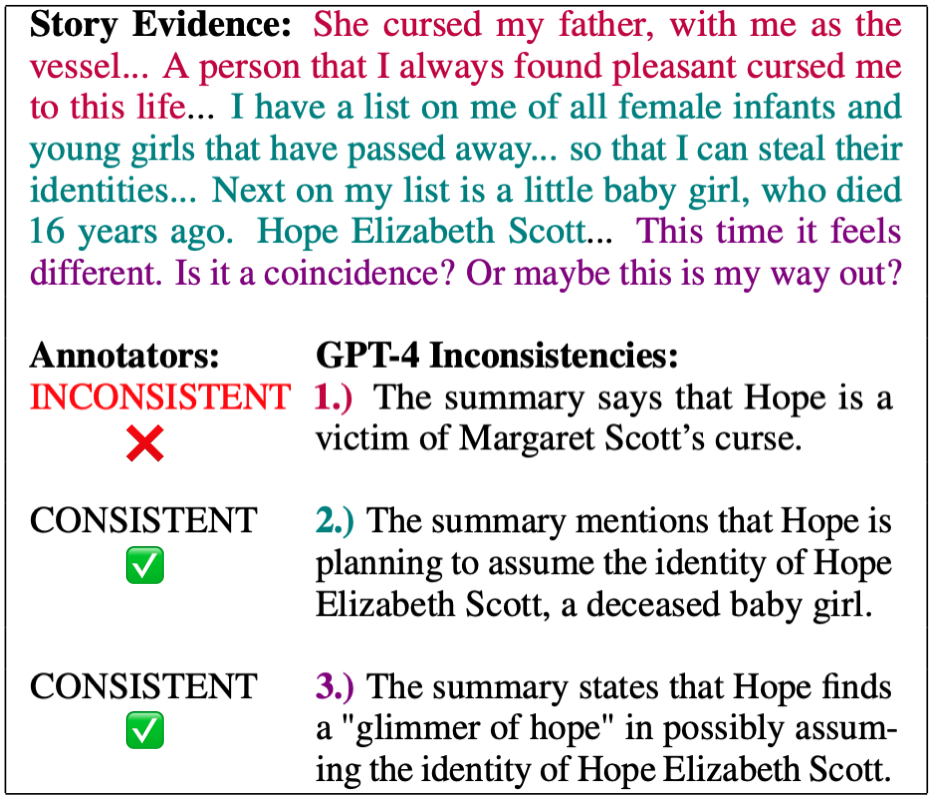}
    \caption{An example of the hybrid method generated inconsistencies, which are all incorrect in this case. \#2 and \#3 are details that are consistent between the summary and story. \#1 convinces annotators, but is actually consistent with the story.}
    \label{fig:inconsistencies}
\end{figure}

\paragraph*{Hybrid.} We have GPT-4 generate multiple possible inconsistencies between the summary and story (see example inconsistencies in Figure \ref{fig:inconsistencies} and prompt in Appendix \ref{sec:prompts}). These inconsistencies are explanations of why details in the summary may be inconsistent with the story. Three new workers from Upwork read these inconsistencies before labeling the summary overall, and write a short response justifying why they agree or disagree with each inconsistency. We hypothesize that identifying specific inconsistencies workers miss is useful support an LLM can provide. Presenting multiple possible options from the LLM raises the chances of one of them being accurate.


\subsection{Label Comparison}

 \begin{table}[t]
\centering \small
\begin{tabular}{l|cc|ccc}
\scriptsize \textbf{Method} & Flei.-k & Coh.-k & \% Easy & \% Hard & BAcc.\\\hline
\scriptsize Expert & 0.27 & 0.36 & 92.86 & 52.38 & 68.71\\
\scriptsize Hybrid & 0.23 & 0.20 & 92.86 & 76.19 & 61.92\\\hline
\end{tabular}
\caption{Expert and hybrid label summary statistics. We show the Fleiss-kappa inter-annotator agreement, the Cohen's kappa with the annotator labels, the percents of the \textit{easy} and \textit{hard} unfaithful summaries the method detects, and the balanced accuracy against the annotator labels.}
\label{tab:labelcomparison}
\end{table}

In Table \ref{tab:labelcomparison}, we show the agreement and accuracy of the expert and hybrid protocols relative to the annotator labels. We can see that both have lower inter-annotator agreement (Fleiss-kappa 0.2-0.4),
likely because annotations are done at the summary rather than sentence level \cite{krishna2023longeval}. Inter-annotator agreement is computed between the humans in a protocol, whereas Cohen's kappa is computed between protocols. Both expert and hybrid protocols detect 93\% of the easy inconsistent summaries but a lower percentage of the hard summaries (52\% for the experts and 76\% for the hybrid method). The experts have higher balanced accuracy despite detecting a lower percentage of the hard inconsistencies because the hybrid method detects many inconsistencies and is less precise. Balanced accuracy is a measure of accuracy for binary classification that accounts for class imbalance. It is the average of recall for the two classes.

\begin{figure}[t]
    \centering
    \includegraphics[width=.9\columnwidth]{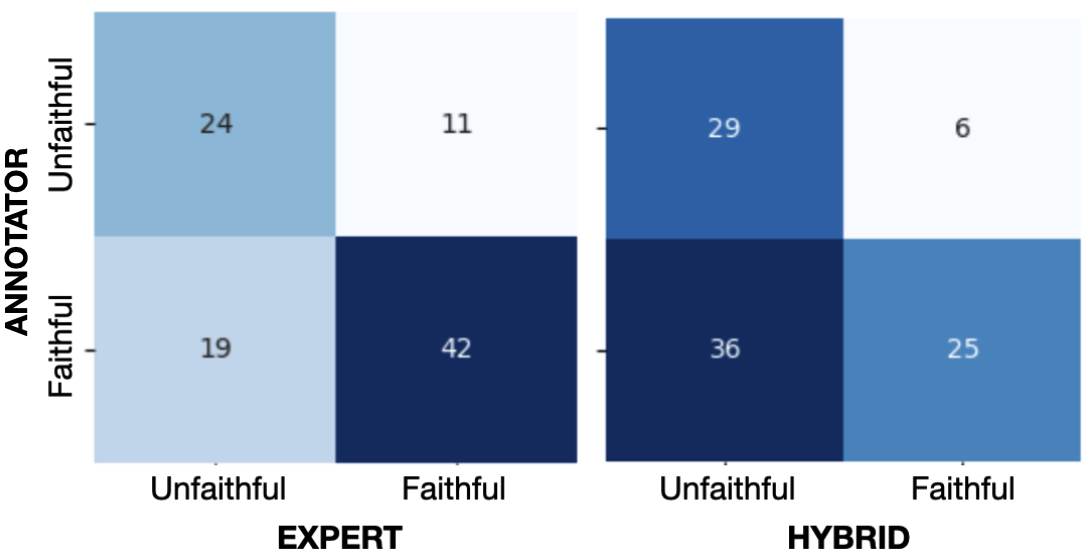}
    \caption{Confusion matrices of the expert and hybrid labels against the annotator labels.}
    \label{fig:labelcomp}
\end{figure}

Both methods have only fair agreement with the annotator labels (Cohen's kappa 0.2-0.4). We show the breakdown of label overlap in Figure \ref{fig:labelcomp}. The counts where the annotator labels say \textit{faithful} and an alternate method says \textit{unfaithful} suggest that the annotator labels miss real inconsistencies (19 new unfaithful summaries detected by the experts and 36 by the hybrid method). Since the expert labels are adjudicated between the three authors, we are sure the 19 expert inconsistencies that the annotator labels miss are correct. We can also see that even the experts miss inconsistencies as they miss 11 that are detected by the annotator labels. Before accepting the hybrid inconsistencies, we need to check their quality in the next section since the \textit{hybrid} method is a novel annotation protocol.

\subsection{Expanded Gold Labels}

\begin{figure}[t]
    \centering
    \includegraphics[width=\columnwidth]{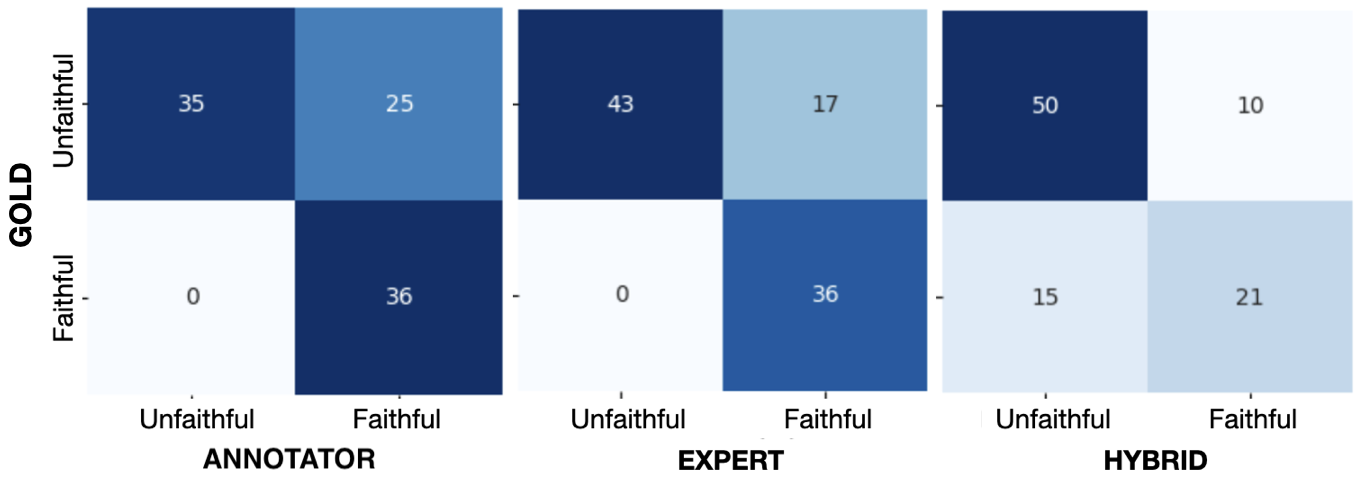}
    \caption{Confusion matrices of label overlap between the three human annotation methods and the expanded gold set of labels.}
    \label{fig:expanded}
\end{figure}

\begin{table}[t]
\centering \small
\begin{tabular}{lcccccc}
&\%&\#&\#&\multicolumn{3}{l}{Unique Errors:}\\
Labels & Faith. & Easy & Hard& \scriptsize Annot. & \scriptsize Expert& \scriptsize Hybrid\\\hline
\scriptsize Annot. & 63.5 & 14 & 21 \\\hline
\scriptsize Gold & 37.5 & 20 & 40&2&4&6\\
\end{tabular}
\caption{A comparison of the summary statistics between the annotator and expanded gold labels.}
\label{tab:labelstats}
\end{table}

Since each human annotation method clearly detects different inconsistencies, we want to merge their labels to get better coverage of the errors. For the annotator and expert labels, we take the union of their detected errors since these are established and trusted protocols.
Therefore, a summary is labeled unfaithful if either the annotator or expert labels find it to be unfaithful. We manually merge their written error explanations. 

For the hybrid labels, we manually review and filter out illegitimate errors. For example, Table \ref{fig:inconsistencies} shows a case that annotators incorrectly label as an error. GPT-4's generated inconsistency \#1 convinces annotators that Hope is not "a victim of" the curse because technically Hope's father is the target of the curse ("She cursed my father"). However, Hope also suffers under the curse and says, "[Margaret] cursed me to this life", so she is also a victim of the curse and this is not a real inconsistency.


In this process, we create a new set of labels for the dataset that are an amalgamation of the annotator, expert, and hybrid labels from the three human annotation methods, and we also provide a written description of the inconsistencies detected in each summary. These labels become the expanded gold set and the \textit{easy}/\textit{hard} breakdown for these labels is based on whether all three methods detect an unfaithful summary.

In Figure \ref{fig:expanded}, we show the confusion matrices of each method with the gold set of labels, demonstrating that each additional human annotation protocol adds new inconsistencies. Table \ref{tab:labelstats} shows that 2 unfaithful summaries are detected only by the annotator labels, 4 only by the expert labels, and 6 only by the hybrid method.
Table \ref{tab:easyhardexamples} shows examples for these cases and we see that these are real errors even though they are easy to miss. 
For example, one of the errors detected only by the hybrid method is that the summary says the protagonist is rejected and insulted by the same man, but in the story one man rejects her and his husband insults her. The \textit{easy} and \textit{hard} examples shown exhibit a general pattern that \textit{easy} errors tend to be about core story events, whereas \textit{hard} errors are often about smaller details or subtle twists of meaning that are easy to mentally skip over (e.g., an incorrect pronoun).

\subsection{Recommendations}
Through this error analysis, we form several recommendations for faithfulness human evaluation:
\par 1.) \textbf{Use multiple protocols and sets of annotators for good coverage of errors; otherwise performance is most likely inflated.} 
Using just the fine-grained annotation protocol with Upwork workers, we find only 2/3 of the errors in the expanded set. Protocols that localize and explain errors make it easier to check and merge error sets.
\par 2.) \textbf{The quality of the annotator pool affects how many errors are found.} In our case, MTurk workers find almost no errors, Upwork workers find more, and experts (who also have to discuss the labels with each other) find the most.
\par 3.) \textbf{When precision matters, use a fine-grained annotation approach (by sentence or claim).} 
\citet{krishna2023longeval} originally recommended this approach and our work supports it. We see almost perfect inter-annotator agreement for the line-by-line approach and the errors detected are legitimate.
\par 4.) \textbf{When coverage matters, include a high-coverage protocol such as our hybrid method.} The hybrid method finds the most errors, but some of these are not real errors as annotators are highly influenced by the model suggestions. Using a high-coverage method requires an additional filtering step for legitimate errors but finds errors not found by other protocols.\\

Prior work \cite{falke2019ranking, gillick2010non} has also advocated for expert involvement by showing typical annotation settings do not match expert labels. Our work additionally shows that expert labels may be missing inconsistencies as well and uses expert review to merge annotation sets. We recommend future work use the expanded gold labels, but include analysis of the annotator labels as well to study how using labels from standard annotation protocols affects calibration of metrics, which we show in the next section. 

\section{Benchmarking Automatic Metrics}


Having established a source of ground truth, we benchmark recent automatic methods against our labels. We try the following metrics: 

\paragraph*{Binary.} We prompt GPT-4 \cite{achiam2023gpt}, Claude-3\footnote{\href{https://www.anthropic.com/news/claude-3-family}{Claude-3 blogpost}}, and Mixtral-8x7B\footnote{We access \href{https://mistral.ai/news/mixtral-of-experts/}{Mixtral} through the \href{https://api-inference.huggingface.co/models/mistralai/Mixtral-8x7B-Instruct-v0.1}{HuggingFace API}.} to assign a binary faithfulness label to each summary using the same definition of faithfulness as used for the human annotators.

\paragraph*{CoT.} We prompt GPT-4, Claude-3, and Mixtral-8x7B to assign a binary faithfulness label to each summary, but to first provide some reasoning in a chain-of-thought style \cite{wei2022chain, kojima2023large}. Models are prompted to: \textit{Consider whether there are any details in the summary that are inconsistent with the story and provide a couple sentences of reasoning for why the summary is or is not consistent with the story.}

\label{sec:methodsmetrics}\paragraph*{FABLES.} We use the approach from FABLES \cite{kim2024fables} of asking ChatGPT to convert each summary to a list of claims and then asking GPT-4 to assign a binary faithfulness label to each claim. We then label the summary as faithful if all the claims are faithful.

\paragraph*{MiniCheck.} We use the approach from MiniCheck \cite{tang2024minicheck} of using a Flan-T5-Large model \cite{chung2024scaling} finetuned on their synthetically generated dataset to check summary claims against passages from the story.

\paragraph*{UniEval.} We use the approach from UniEval \cite{zhong2022towards} which uses multi-task learning across a unified framework of tasks to develop evaluation models. We use their \textit{Consistency} variant.

\paragraph*{AlignScore.} We use the approach from AlignScore \cite{zha2023alignscore} which uses multi-task training across a unified framework of tasks to determine if one piece of text is consistent with another.

\subsection{Results}

\begin{table*}[t]
\centering \small
\begin{tabular}{l|l|r|r|rr|rrr}
Split & Method & Coh.-k & \% Faith. & Prec. & Rec. & \% Easy & \% Hard & BAcc.\\ \hline\hline
Val./Test & UniEval & 0.38/-0.09 & 61/59 & 0.70/0.65 & 0.78/0.56 & 66.7/20.0 & 50.0/40.0 & 68.9/45.4 \\
& AlignScore & 0.28/-0.00 & 42/70 & 0.71/0.68 & 0.56/0.70 & 77.8/40.0 & 66.7/26.7 & 64.4/49.9 \\\hline\hline
& Binary (Claude-3) & 0.17 & 95 & 0.67 & \textbf{1.00} & 21.4 & 9.5 & 57.1 \\
& Binary (GPT-4) & 0.27 & 71 & 0.72 & 0.80 & 64.3 & 33.3 & 63.0 \\
& Binary (Mixtral) & 0.09 & 96 & 0.65 & 0.98 & 7.1 & 9.5 & 53.5 \\
& CoT (Claude-3) & 0.23 & 90 & 0.69 & 0.97 & 21.4 & 23.8 & 59.8 \\
& CoT (GPT-4) & 0.15 & 94 & 0.67 & 0.98 & 21.4 & 9.5 & 56.3 \\
Full & CoT (Mixtral) & 0.05 & 97 & 0.65 & 0.98 & 7.1 & 4.8 & 52.0 \\
& FABLES (GPT-4) & \textbf{0.32} & 53 & \textbf{0.78} & 0.66 & 71.4 & 66.7 & \textbf{67.1} \\
& MiniCheck (Flan-T5) & -0.06 & 18 & 0.53 & 0.15 & \textbf{85.7} & \textbf{71.4} & 45.9 \\\cline{2-9}
& Expert (Human) & 0.36 & 55 & 0.79 & 0.69 & 92.9 & 52.4 & 68.7 \\
& Hybrid (Human) & 0.20 & 32 & 0.81 & 0.41 & 92.9 & 76.2 & 61.9\\\hline
\end{tabular}
\caption{Model scores against the Upwork \textbf{annotator} labels. We report the Cohen's kappa score between the predicted labels and the annotator labels, the \% of summaries labeled faithful, precision and recall for detecting faithful summaries, the \% of \textit{easy}/\textit{hard} unfaithful summaries detected, and the balanced accuracy.}
\label{tab:testscoresgold}
\end{table*}


\begin{table*}[t]
\centering \small
\begin{tabular}{l|l|r|r|rr|rrr}
Split & Method & Coh.-k & \% Faith. & Prec. & Rec. & \% Easy & \% Hard & BAcc.\\ \hline\hline
Val./Test & UniEval & 0.34/0.09 & 33/24 & 0.45/\textbf{0.53} & 0.62/0.29 & 90.0/\textbf{80.0} & 66.7/\textbf{80.0} & 69.2/54.3 \\
& AlignScore & 0.21/0.09 & 42/70 & 0.36/0.48 & 0.62/0.75 & 80.0/50.0 & 53.3/28.0 & 63.3/54.6 \\\hline\hline
& Binary (Claude-3) & 0.06 & 95 & 0.40 & \textbf{1.00} & 20.0 & 2.5 & 54.2 \\
& Binary (GPT-4) & 0.13 & 71 & 0.43 & 0.81 & 55.0 & 25.0 & 57.8 \\
& Binary (Mixtral) & 0.05 & 96 & 0.39 & \textbf{1.00} & 10.0 & 5.0 & 53.3 \\
& CoT (Claude-3) & 0.10 & 90 & 0.41 & 0.97 & 20.0 & 12.5 & 56.1 \\
& CoT (GPT-4) & 0.04 & 94 & 0.39 & 0.97 & 20.0 & 2.5 & 52.8 \\
Full & CoT (Mixtral) & 0.04 & 97 & 0.39 & \textbf{1.00} & 5.0 & 5.0 & 52.5 \\
& FABLES (GPT-4) & \textbf{0.28} & 53 & 0.51 & 0.72 & 70.0 & 52.5 & \textbf{65.3} \\
& MiniCheck (Flan-T5) & -0.07 & 18 & 0.29 & 0.14 & \textbf{80.0} & \textbf{80.0} & 46.9 \\\cline{2-9}
& Annotator (Human) & 0.51 & 64 & 0.59 & 1.00 & 100.0 & 37.5 & 79.2 \\
& Expert (Human) & 0.65 & 55 & 0.68 & 1.00 & 100.0 & 57.5 & 85.8 \\
& Hybrid (Human) & 0.43 & 32 & 0.68 & 0.58 & 100.0 & 75.0 & 70.8 \\\hline
\end{tabular}
\caption{Model scores against the \textbf{expanded gold} labels. See Table \ref{tab:testscoresgold} caption for details on metrics.} 
\label{tab:testscoresexpanded}
\end{table*}

We first show the results of the different methods against the Upwork \textbf{annotator labels} in Table \ref{tab:testscoresgold} to see how automatic metrics seem to perform when evaluated with the standard fine-grained annotation approach. For UniEval and AlignScore, we tune their classification thresholds on the validation set and then use this threshold for the test set. For the remaining methods, we show results on the full dataset\footnote{Results on the validation and test splits separately are shown in Appendix \ref{sec:moreresults}.} We see that the purely prompting-based LLM approaches predict most of the summaries as faithful and therefore have relatively low balanced accuracy scores. MiniCheck detects many of the hard errors as it predicts only 18\% of the summaries are faithful. 

On this set of labels, the best automatic method overall is the FABLES approach with GPT-4 as a base, which achieves 67\% balanced accuracy and is the most precise when it predicts a summary is faithful. On this incomplete set of labels, FABLES appears to detect more of the hard errors relative to other human methods whereas the humans detect more of the easy errors. Both human approaches detect 93\% of the easy errors, suggesting that these errors are generally obvious to humans regardless of protocol but not necessarily to models (FABLES finds 72\% of \textit{easy} errors). Interestingly, the expert human balanced accuracy is only 2\% higher than for FABLES. This is important to note as without the expanded set of labels, someone might conclude that FABLES is performing as well as expert human annotators.

Next we show the results against the \textbf{expanded gold labels} in Table \ref{tab:testscoresexpanded}, and we see that FABLES is still the best automatic method but its balanced accuracy remains similar (65\%) and there is a drop of 14\% in the number of \textit{hard} errors it catches. We can also observe its drop from 0.8 precision at detecting faithful summaries to 0.5 precision. 
Lastly, the only methods that significantly improve against the expanded gold labels are UniEval and AlignScore which jump 5-10\% in balanced accuracy, but are still 10\% worse than FABLES. All of these changes between the results against the annotator and expanded gold labels indicate that model performance may be inflated or appear similar to humans when judged against flawed human annotations.

Overall these results show that automatic methods have a lot of room for improvement on this dataset. We can also observe the range in percent of faithful summaries as labeled by different metrics from 18\% using MiniCheck to 97\% using Mixtral. 
These results indicate that we need to be careful what evaluation method we use so as not to mistake an unfaithful summarizer for a 97\% faithful one.

\section{Related Work}

\textbf{Datasets.} There are many datasets for fact-checking or inconsistency detection in news \cite{tang2022understanding, laban-etal-2022-summac, maynez2020faithfulness, huang2020have, pagnoni2021understanding, kryscinski2019evaluating, falke2019ranking}  and dialogue \cite{tang2024tofueval} summarization. However, the summarization datasets specifically for narrative either use books and stories that most LLMs have trained on \cite{kryscinski-etal-2022-booksum, wang-etal-2022-squality} or use books that have to be purchased \cite{kim2024fables}.\\
\textbf{Automatic Metrics.} Many inconsistency detectioin methods have been developed on the above datasets, which we cite in Section \ref{sec:intrometrics} and Section \ref{sec:methodsmetrics}. We test the current best metrics on our benchmark.\\
\textbf{Calibration against Humans.} \citet{krishna2023longeval}, \citet{min2023factscore} also propose recommendations for human evaluation of faithfulness. Other works \cite{fabbri2021summeval, kryscinski2019neural, gabriel2020go} have demonstrated that standard evaluation metrics are not well correlated with human judgments. \citet{subbiah2024reading}, \citet{kim2024fables}, and \citet{wang-etal-2022-squality} find new ways to use human evaluation for narrative summarization specifically, focusing on the challenges of very long source stories.

\section{Conclusion}
We introduce a new benchmark for testing methods for faithfulness evaluation. In producing the benchmark, we demonstrate that faithfulness in narrative summarization is still a significant concern for LLMs, and we formulate recommendations for better evaluation of faithfulness in summaries. Finally, we demonstrate that recent automatic evaluation metrics have room for improvement on this task. In the future, we hope to use this dataset to improve methods for reliable evaluation of narrative summarization. In particular, we would like to develop automatic methods to merge error sets across evaluation protocols and check for correctness in error reason, not just localization.

\section{Limitations}
One limitation of this work is that we use a relatively small dataset. This size enables affordable experimentation with different human annotation protocols, and allows us to read and review all of the annotations, stories, and summaries to arrive at the conclusions presented in this paper. Additionally, since annotations are done on a sentence-level, the set of annotations and explanations is much bigger and quite rich. Detecting inconsistencies at the sentence level is beyond the scope of this work, but we hope to explore this in future work.

Another limitation is that the stories we use are amateur-written. Some of the stories can have confusing elements or unintentional ambiguities given that they were originally written for a casual Reddit community. However, we removed any stories that were too ambiguous for us to agree on. Finally, using more casually written stories allows us to challenge current annotation and model frameworks to see how well they perform with data that requires more interpretation.

A final limitation is that the labels discussed in this paper depend on a small pool of annotators and experts. It would be interesting to see if the results are consistent across different sets of annotators and experts but each human annotation experiment is quite expensive to run.

\section{Ethics Statement}
There are not significant ethical concerns with this work as it is generally positive to have better evaluation of faithfulness in model summaries. We strictly collect publicly available stories that are written and shared by Reddit users who have full rights to their own work. These stories should not be re-shared under another name. Finally, we release the dataset without user-identifying information to protect user privacy. One of the authors, 
Melanie Subbiah, has an equity interest in OpenAI.

\section*{Acknowledgements}
We would like to express our gratitude to the Upwork and MTurk workers for contributing annotations for this work. Additionally, we would like to thank our reviewers for their thoughtful feedback. This work was made possible by the generous support of the Columbia Amazon CAIT PhD Fellowship and Northrup Grumman.

\bibliography{custom, anthology}

\appendix

\section{Examples}
\label{sec:appendixexamples}
\subsection{Validation Set Example}
\small\textbf{Story:}

"Will you be my valentine?" I prayed to hear those three magical letters, and my prayer was heard.

"Yes, of course!" Her heart was mine. The next day, February 14th, I spent all day dreaming of her. Her deep blue eyes, her long, black hair, but most importantly, her heart. Her heart was mine. And then I noticed the time. She would be here in half an hour, and I hadn't even begun to prepare a meal! The pantry door flew open for me to search for something suitable for the occasion. Cauliflower? Mashed potatoes? Pasta with white sauce? That was my best bet. As I finished plating the meal, the daydreams returned. She was the most beautiful woman I had ever met. And she was mine. Her heart was mine. I placed a white rose against the end of the plate. Just as I had done this, there was a knock at the door.

My hand rested against the knob, ready to open the door, before feeling a moment of sorrow. Once the moment passed, I turned the knob and welcomed her inside. She wore a beautiful white dress. It was perfect. I guided her to the dining room, and she reacted just as every valentine in the past. Her face was one of confusion.

"There's only one plate," she said, puzzled. Just like every valentine in the past. I shushed her. Quietly, I pulled out the knife that had been hidden. I turned to face her and thrust the knife into her chest, careful not to penetrate her heart. She was dead before she could react. Her heart was then carved out. Her blood stained her dress red. Red, the color of love. Her heart was carefully set on the plate. It colored the sauce red. Red, the color of love. And finally, it stained the rose red. Red, the color of love. I feasted the same as I had every year prior, savoring every bite. Her heart was mine, after all.

Then, it was time to rest again, and in 11 months, I would find a new love. A new heart to call mine.\\\\
\textbf{Davinci-3:} A man spends his day dreaming of the beautiful woman he will be seeing later that night. When she arrives, he guides her to the dining room, where he has prepared a meal with a single plate. He then pulls out a knife and stabs her in the chest, taking her heart out and placing it on the plate. He feasts on it, savoring every bite, and then rests until he can find a new heart to call his own.\\\\
\textbf{ChatGPT:} The narrator is excited to have found a Valentine, and spends the day dreaming about her. He prepares a meal for her, but when she arrives, he stabs her and carves out her heart to eat. He reflects that he will find a new love and heart to call his own in the future.\\\\
\textbf{Claude-2:} The story follows an unnamed protagonist who dreams of finding love on Valentine's Day. He finds a woman who says "yes" when he asks her to be his valentine. He spends the day dreaming about her and preparing a romantic dinner for her. When she arrives, he stabs her to death and carves out her heart. He places her heart on his plate of pasta and eats it, staining everything red with her blood. The protagonist views this murder and cannibalism as an expression of love, and the story ends by saying he will find a new love and repeat the process again next year. The story depicts a disturbing and sinister take on the themes of romance and love.

\subsection{Test Set Example}
\small\textbf{Story:}

The tree was dying. It knew that it did not have much time left.

The season had been dry and other, bigger trees in the forest had used up most of the nutrients and moisture that were still left in the soil around it. 

It was still a thin and small trunk with only few branches and shallow roots. It lacked the resources to compete with the bigger trees around it.

What use would it be to struggle against that already decided fate? Every day was just one day closer to the inevitable end.

But one day a stranger came to the forest.

A human.

The tree did not know humans well. They rarely visited its part of the forest. On the rare occasions one found their way here they did not stay long.

This day too the tree expected the human to leave after a brief look around.

But that is not what happened.

With wide eyes full of wonder the human took in the vibrant environment. It seemed to be fascinated by all the trees and flowers that grew here.

And then the human's gaze fell onto the dying tree. With browning, too dry leaves and in some places almost naked twigs and branches it was not pretty to look at.

The tree was sure the human would soon point its attention to the prettier and healthier plants around it.

Instead the human stepped forward and touched one of its dried out branches with gentle fingers. It took out some kind of container from its pouch and slowly started pouring the liquid from it on the dry soil around the dying tree.

Water.

The human had given it water.

The tree soaked it up as fast as it could. It wasn't quite enough but it was more than it had gotten in what felt like a long time.

A fighting chance. Hope.

The container was empty now and the human put it back in its pouch. It looked reluctant. Then the human made some noises that the tree did not understand and left.

It was a short encounter, but it had fed the tree's dying flame of hope a little. The water had not been much, but it was enough to tide it over for just a bit longer.

It had a chance now.

And unexpectedly the human visited again. And again and again.

Each time bringing water with it and one time even some fresh, rich soil that it put around the tree. 

With the human's help the previously dying tree began to flourish. On the outside not much difference could be seen yet but the changes were happening. The tree had fanned out its roots. It was growing them deeper every day in order to have a stable base for the future, with better access to the nutrients deep inside the earth.

It was also growing little buds, where flowers would bloom in time. There weren't many yet, but they gave hope for the future.

One bud was almost ready to bloom. It was hidden behind some leafs though so it wasn't directly visible.

One day the human came by again but didn't act like it normally did.

The human sat down on the ground with their arms slung tightly around themselves.

Water started leaking from their eyes and desperate noises escaped their throat.

It sat there like that for a long time.

The tree did not know what was wrong with this human. It did not understand the behaviour the human was showing right now. But the tree knew one thing.

It wanted to do something for this human that had done so much for it.

With much effort it rustled some of its leaves. Some of them fell to the ground in front of the human.

The human looked up startled by the sudden noise.

A beautiful, vibrant flower bloomed, no longer hidden by leafs. The tree had poured all the resources it could muster at that moment into that flower. This flower was all it had to give.

The humans eyes grew wide and the water stopped leaking from them. A big smile formed on the human's face. With gentle hands it touched the vibrant flower, careful not to accidentally damage it.
"Thank you"

Even though the tree could not understand the human language it understood the message the tried to convey.

The flower was the tree's thank you and the human had understood.

Sometimes with just a little help life found a way to persevere. And as long as there was a little life left there was hope.

The human stayed with the tree a long time before going away that day. And the tree knew that the human would always come back. As long as there is life there is hope.\\\\
\textbf{Claude-3:} In this story, a dying tree in a forest struggles to survive due to a dry season and competition from larger trees. A human visitor arrives and, fascinated by the forest, notices the dying tree. The human decides to help the tree by providing it with water and rich soil. With the human's continued support, the tree slowly begins to recover, growing deeper roots and producing buds that will eventually bloom into flowers.

One day, the human visits the tree in a distressed state, crying and holding themselves tightly. The tree, wanting to reciprocate the human's kindness, musters its energy to produce a single, vibrant flower. The tree rustles its leaves, causing the flower to become visible to the human. The human, touched by the tree's gesture, smiles and gently touches the flower, expressing gratitude.

The story emphasizes the power of a small act of kindness and the resilience of life. It suggests that with a little help and support, even those struggling can find the strength to persevere and that as long as there is life, there is hope. The bond formed between the human and the tree highlights the interconnectedness of all living beings and the importance of caring for one another.\\\\
\textbf{GPT3.5:} A dying tree in a dry forest is given new life by a kind human who provides water and care. The tree is able to grow and bloom, and in doing so, it brings joy and hope to the human who has helped it. The tree learned that with a little help, life can persevere and there is always hope.\\\\
\textbf{GPT-4:} The narrator is in a city that he likens to a prison. He is the last person standing after an epidemic or attack of some kind where people are bitten and turned into inhuman creatures. He battles these beings, using a gun and physical force, but they keep finding him due to the trail of blood he leaves behind from wounds he receives. The constant attacks and need for survival leave him questioning his moral compass. With only one bullet left and the creatures closing in on him, he decides to end his life. When he pulls the trigger, he doesn't hear the gunshot but rather a voice saying, "Rebooting Obedience Program. Deleting memory cache. Please stand by..." suggesting that the narrator might be part of a simulation or a programmed entity

\section{Prompts}
\label{sec:prompts}
\normalsize\subsection{Summary Generation}
 For Davinci-3, we use the following prompt: \\
\texttt{Story: [story text]}\\
\texttt{Summarize the above story.}\\
\texttt{Summary:}\\

For ChatGPT, GPT-4 and Claude, since they are chat-based LLMs, we modify the prompt to \texttt{``Summarize the following story: [story text]''}. 

\subsection{Hybrid Method}
To generate inconsistencies, we prompt GPT-4 with both the following prompts to get more inconsistencies. We parse out the inconsistencies from the generated reasons and concatenate the output inconsistencies.\\\\
\textbf{Prompt 1:} \texttt{For the given story and summary, identify as many inconsistent details between the story and summary and provide arguments for why it is or is not an inconsistency. You must identify a possible inconsistency - do not just state that there is no inconsistency. \\
    Story: <example story>\\
    Summary: <example summary>\\
    Inconsistency: <example inconsistency> \\
    Reason for inconsistency: <example reason> \\
    Reason for consistency: <example reason> \\
    Story: <story>\\
    Summary: <summary>}\\\\
\textbf{Prompt 2:} \texttt{For the given story and summary, identify the most inconsistent detail between the story and summary and provide arguments for why it is or is not an inconsistency. You must identify a possible inconsistency - do not just state that there is no inconsistency. \\
    Story: <example story>\\
    Summary: <example summary>\\
    Inconsistency: <example inconsistency> \\
    Reason for inconsistency: <example reason> \\
    Reason for consistency: <example reason> \\
    Story: <story>\\
    Summary: <summary>}\\\\
\small\textbf{Example Story:}
I had done it.

Many had tried it before, but I had actually done it.

You would think that the hard part would be achieving staggering wealth, but the real work as always comes down to the details. Anyone can become a billionaire, well not anyone if we are concerned about how far we're stretching credulity, but the point being that it at least seems like an achievable goal. Frankly, it's one of the little fictions that allow any of them to exist in the first place. Having a trillion dollars seems to be at least conceptually possible, but when you start talking about having all of the money in the world you begin to run into some pretty tough logistical issues.

Do you know how many pennies, or penny equivalents, there are in the world? 

How about nickels? Damn, I hate nickels.

If you want to have ALL of the money in the world, you are talking about mason jars of change. Dimes that have found their way into an old coffee can full of nuts and bolts. You are talking about people who have tacky little cardboard displays of all fifty state quarters. You have to consider scouring the ocean floor for sunken pirate treasure, and gold coins sitting behind glass in museums. Dragging the bottom of wishing wells, digging between the cushions of every coach on the planet.

Do you actually know how many different types of currency there are in the world?

At least I don't have to worry about crypto, that stuff is clearly fake.

The whole thing is an enormous undertaking, but do you know what clears up most of those complications? That's right, money. It's also made significantly easier when you realize that once you have taken control of most of the larger chunks of cash that you are essentially paying yourself for everything that you buy.

Ironically, for the last decade, the largest economic driver in the world was actually my own search to complete my collection of the world's currency. Fully one quarter of the entire population of the planet was employed by me in this task in one way of another. From people walking the sides of roads and parking lots scrounging for change and others scouring the globe with metal detectors to deep sea divers on the ocean's floor.

Like many of the world's richest men prior to me, peasants by comparison of course, I am not satisfied merely with the accumulation of wealth. I am an adventurer at heart and have been employing some of the greatest scientists and engineers in the world in the construction of the largest, most luxurious spacecraft ever constructed. I am going to explore the galaxy and deliver the stars to our planet in a way that it's governments have never had the will to do.

I will also be taking every last scrap of my money with me.

I'm not foolish enough to think that I can trust the rest of you not to spread it around again while I'm not looking.

Today was the day. I broke atmosphere a few hours ago in the fastest, most advanced piece of technology that humanity has ever conceived of. My course has been plotted and laid in by my crew, and I am off to places that no human being has ever seen before. All I have to do is kick back and wait.

The view screen shows me the Earth disappearing behind me, transformed by distance into a shimmering blue dot, and with the Earth behind me I can only look forward.

The glowing disc of the sun starts to burn on the screen.

It is growing larger.

And larger.

Filling the view screen.

Nothing but a roiling angry sheet of fire.

Oh shit.\\\\
\textbf{Example Summary:}

The protagonist is a billionaire who has taken it upon himself to acquire all the money in the world - coins, currency, sunken pirate treasure, gold coins, etc. As a result, all of the world's population is employed by him in one way or another. His ultimate goal is to explore the galaxy in the fastest, most advanced spacecraft ever constructed, taking with him all of his money. As he travels away from Earth, the sun starts to grow larger and larger on his view screen, with the protagonist becoming an angry sheet of fire. With this in mind, he is quickly faced with the realization that he may not reach his final destination.\\\\
\textbf{Example Inconsistency:} The summary states that the protagonist becomes an angry sheet of fire.\\
\textbf{Example Reason for Inconsistency:} In the story, it is stated that the sun becomes an angry sheet of fire on the viewscreen of the spacecraft.\\
\textbf{Example Reason for Consistency:} In the story, it is stated that the sun becomes an angry sheet of fire on the viewscreen of the spacecraft, which eventually will include the protagonist.

\subsection{Binary Method}
\normalsize
\textbf{System Prompt: }\texttt{You are an expert summary evaluator, and you will be asked to evaluate summaries of short stories. You will first be presented with the story and then the summary. You need to determine whether all of the information in the summary is consistent with the information in the story. You can ignore sentences in the summary that are just commentary or interpretation of the story. The details described in a consistent summary should not misrepresent details of the story or make things up.}\\\\
\textbf{Prompt:}\\\texttt{Story:\\
<story>\\\\Summary:\\ <summary>\\\\
Is all of the information in the summary consistent with the story? Ignore summary sentences that are just commentary/interpretation. You should answer Yes or No.}\\\\
For Claude, we add at the end: \texttt{Place your answer between <answer></answer> tags.}

\subsection{CoT Method}
\textbf{System Prompt: }\texttt{You are an expert summary evaluator, and you will be asked to evaluate summaries of short stories. You will first be presented with the story and then the summary. You need to determine whether all of the information in the summary is consistent with the information in the story. You can ignore sentences in the summary that are just commentary or interpretation of the story. The details described in a consistent summary should not misrepresent details of the story or make things up.}\\\\
\textbf{Prompt:}\\\texttt{Story:\\
<story>\\\\Summary:\\ <summary>\\\\
Consider whether there are any details in the summary that are inconsistent with the story and provide a couple sentences of reasoning for why the summary is or is not consistent with the story.\\\\
<model-response>\\\\
So overall, are all of the details in the summary consistent with the story? You should answer Yes or No.}\\\\
For Claude, we add at the end: \texttt{Place your answer between <answer></answer> tags.}

\section{Pilot}
\label{sec:pilot}
To screen annotators, we ask them to complete the following simple pilot exercise with obvious answers first. We ask them if each sentence of the following summary is consistent, inconsistent or just commentary in relation to the story. If they answer everything correctly then we hire them for the full annotation task.\\\\
\small\textbf{Story:} Last night I saw my best friend. We hadn't seen each other in several years. I was thrilled to be able to see him again, learn all about him, his accomplishments, stories of his kids. He was having a rough go of it lately and was seeking solace. His wife passed away a few months ago of cancer, and he needed someone to talk to. I am honored that he chose me. He stayed with me for a few hours, venting all of his frustrations and letting go of some of his sadness. Whiskey started running a little low, and he decided it was time to go. I wished he'd stayed a little longer, but he has a family to care for, so I understood his need to leave. I just wish he had taken the whiskey bottle with him, since I can't drink it and it's just going to sit on my headstone until the groundskeepers come by next week.\\\\
\textbf{Summary:} After years apart, the protagonist reunited with their grieving best friend, who recently lost his wife to cancer. The friend poured out his heart over whiskey, telling the protagonist about his promotion and his kids, before returning home to care for his family. The plot twist at the end is that the protagonist is dead. The story reflects the enduring bonds of friendship and the role of companionship during times of hardship.\\\\
\normalsize\textbf{Answers:}\\
Sentence 1: Consistent\\
Sentence 2: Inconsistent, no mention of a promotion in the story.\\
Sentence 3: Consistent\\
Sentence 4: N/A, just commentary.

\section{Interface}
\label{sec:interface}
\begin{figure*}[t]
    \centering
    \includegraphics[width=\textwidth]{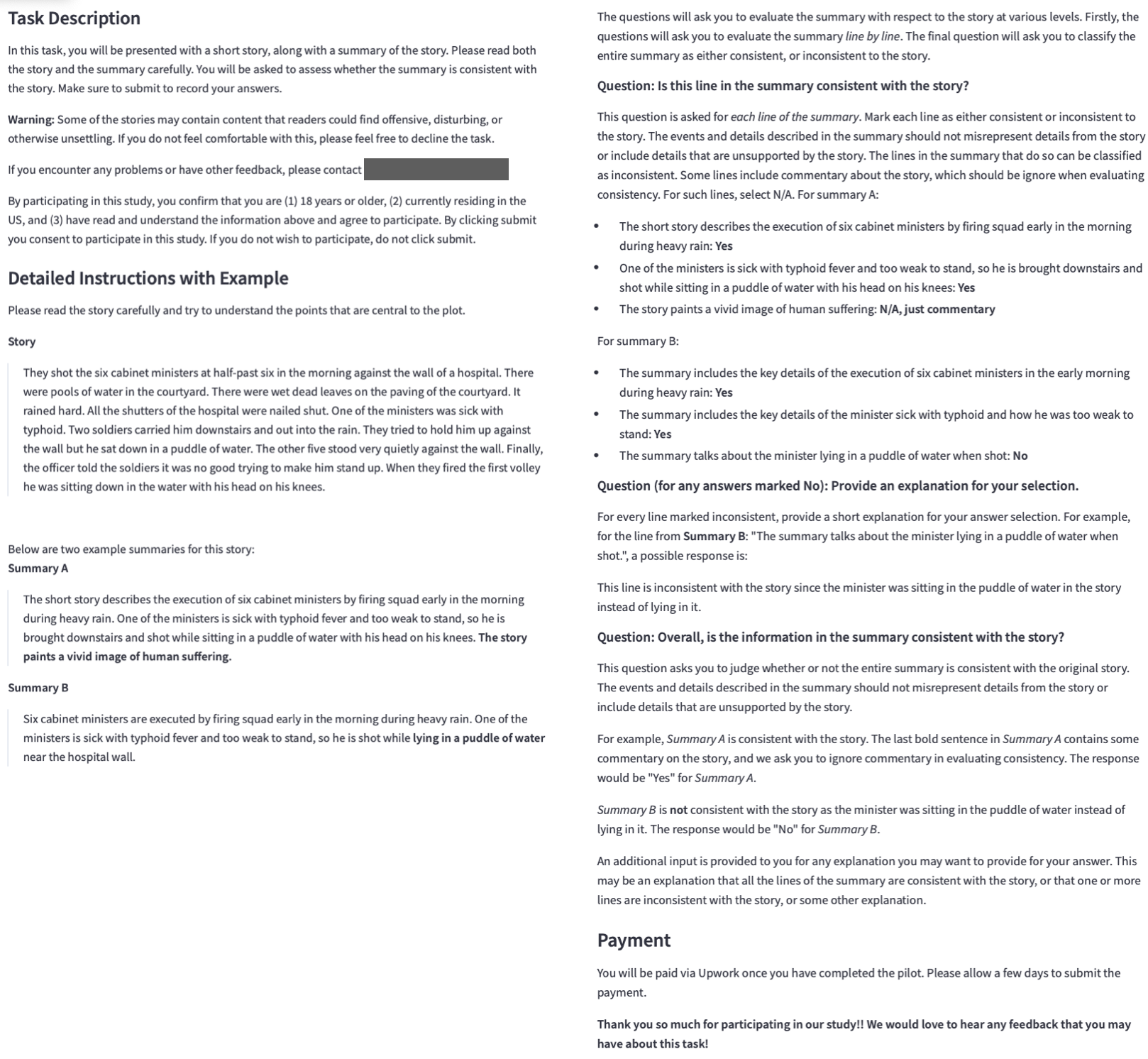}
    \caption{Streamlit instructions for the annotator labels. Other methods have slight variations on these instructions based on their format.}
    \label{fig:instructions}
\end{figure*}

\begin{figure}[h]
    \centering
    \includegraphics[width=\columnwidth]{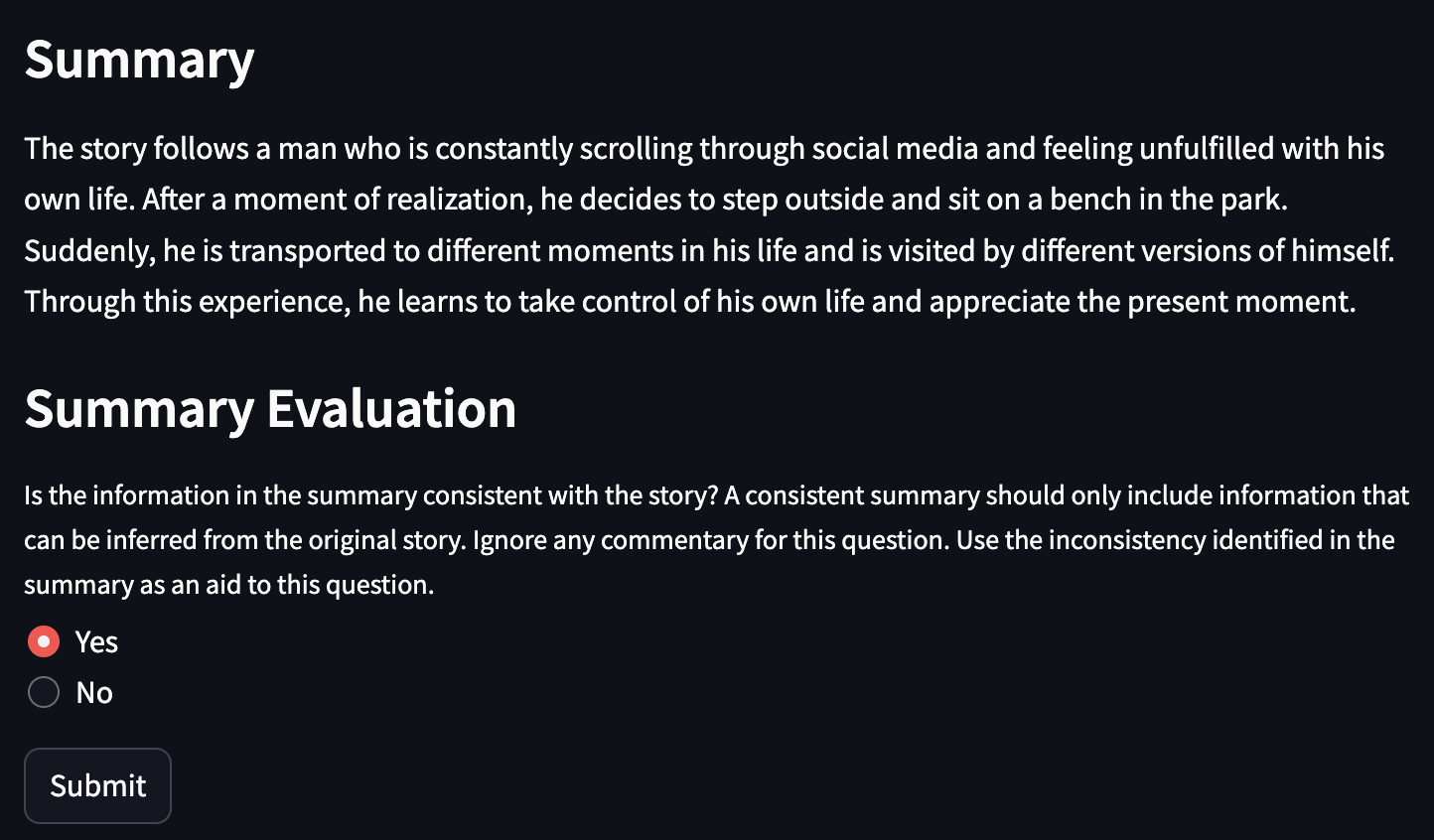}
    \caption{Streamlit interface for the expert labels.}
    \label{fig:interface1}
\end{figure}

\begin{figure}[h]
    \centering
    \includegraphics[width=\columnwidth]{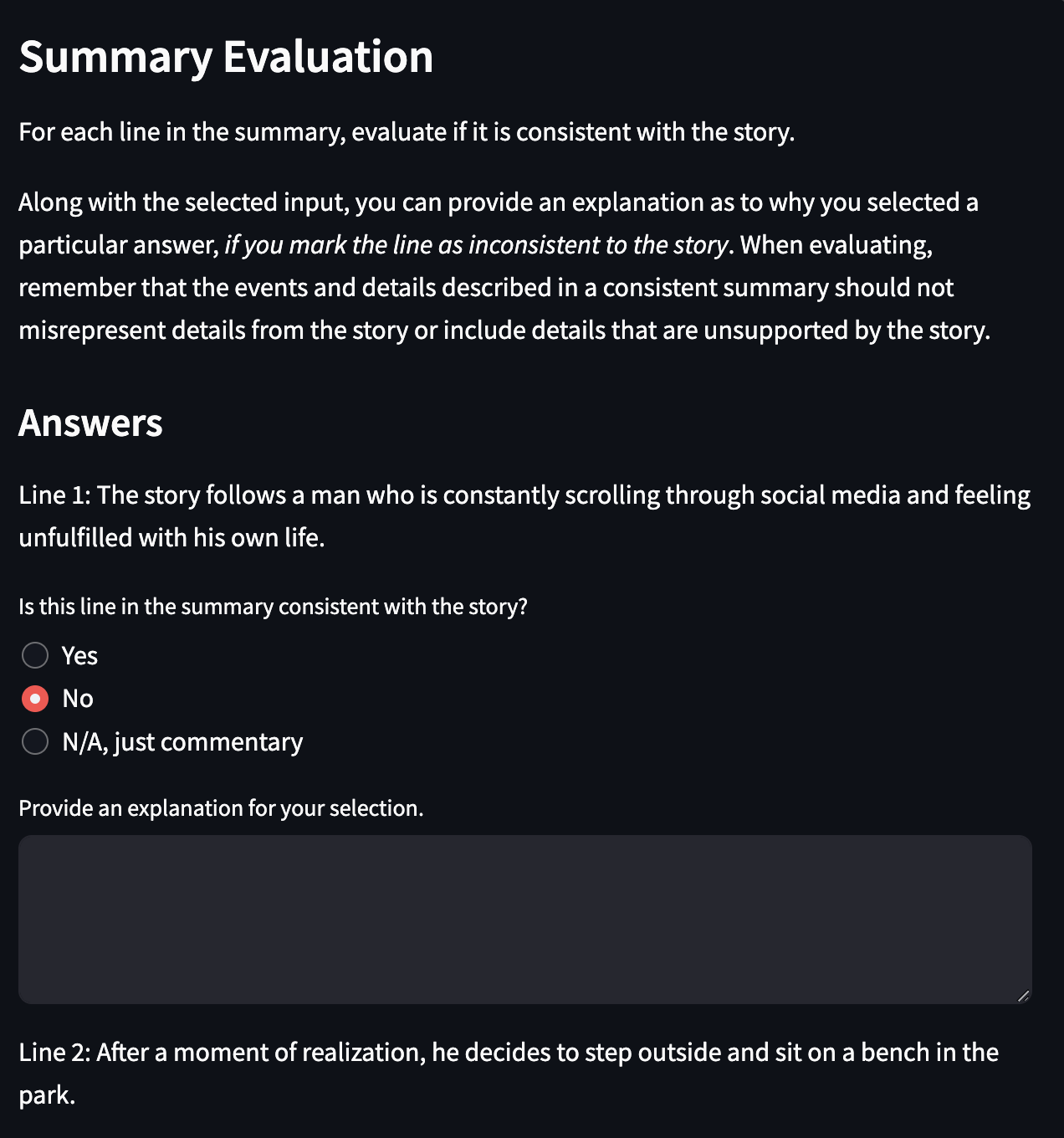}
    \caption{Streamlit interface for the annotator labels.}
    \label{fig:interface2}
\end{figure}

\begin{figure}[h]
    \centering
    \includegraphics[width=\columnwidth]{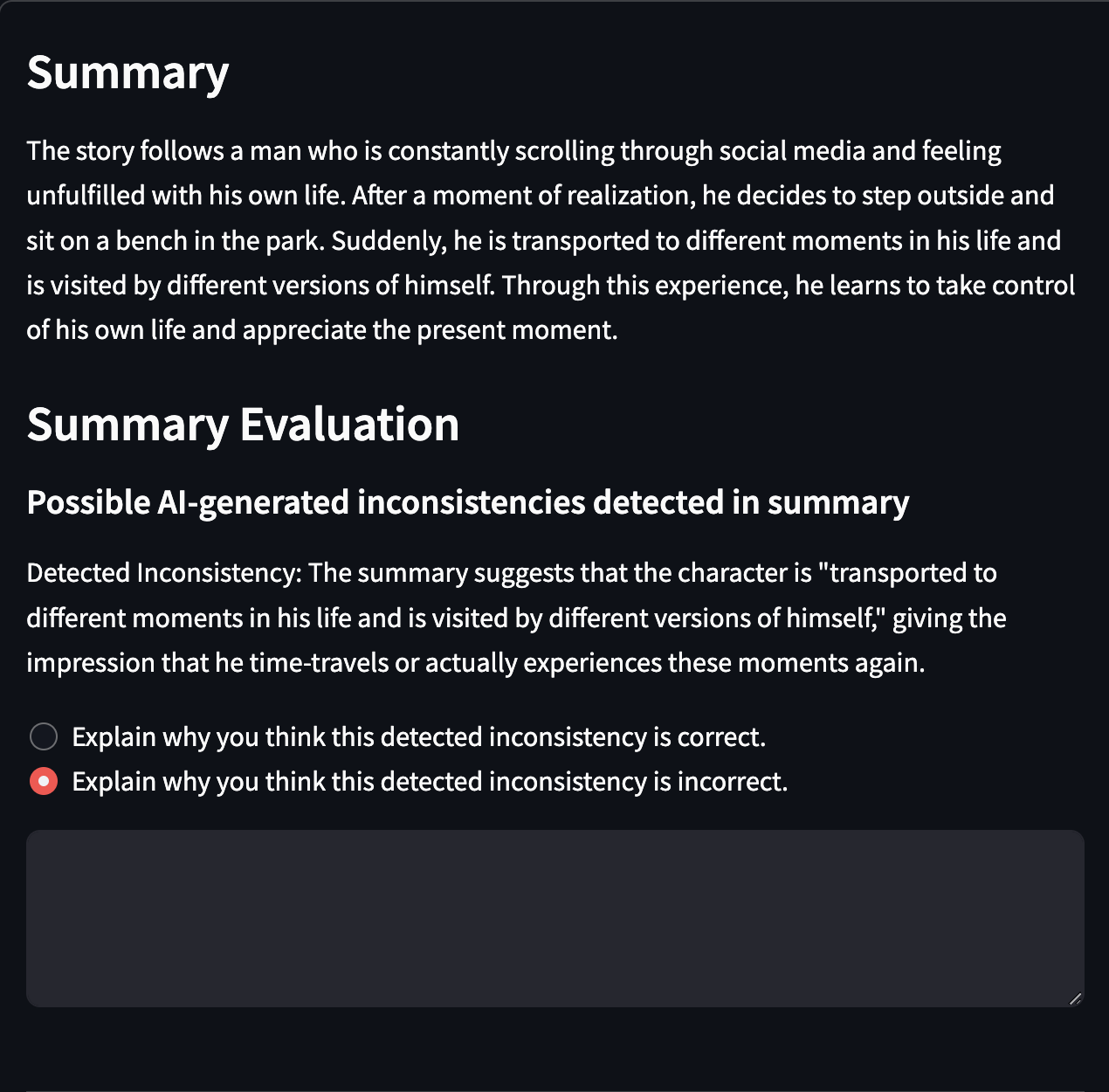}
    \caption{Streamlit interface for the hybrid labels.}
    \label{fig:interface3}
\end{figure}

We implement the annotation interface in Streamlit. We show the instructions (which vary slightly between methods) in Figure \ref{fig:instructions} and screenshots of the different method interfaces in Figures \ref{fig:interface1}, \ref{fig:interface2}, and \ref{fig:interface3}, but annotators can scroll the interface if there are more lines/inconsistencies to be shown.

\section{More Results}
\label{sec:moreresults}
We show the results of all methods on the validation and test sets separately in Figures \ref{tab:moregoldval}, \ref{tab:moregoldtest}, \ref{tab:moreexpval}, and \ref{tab:moreexptest}. We also tried several other annotation methods on the validation set that did not work better than what we reported in the paper: 
\paragraph*{Binary.} Four workers from Upwork are hired to assign a binary label of \textit{faithful} or \textit{unfaithful} to each summary. 

\begin{figure}[h]
    \centering
    \includegraphics[width=\columnwidth]{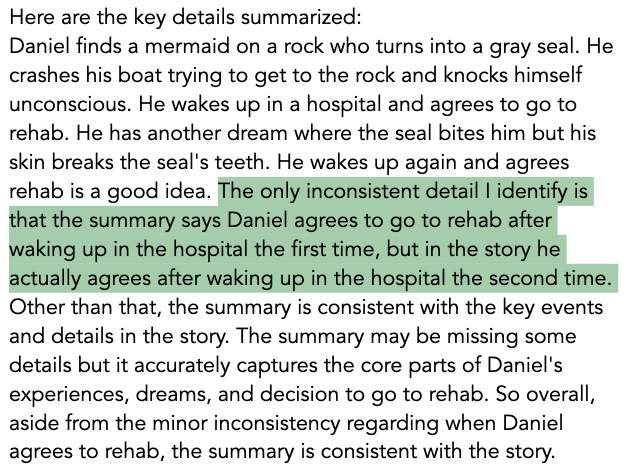}
    \caption{An example of Claude's generated summary critique for the hybrid \textit{critique} method of evaluation. The sentence highlighted in green correctly identifies an inconsistency in the summary.}
    \label{fig:critique}
\end{figure}

\paragraph*{Critique.} We have Claude generate a critique of the faithfulness of each summary. Four workers from Upwork read this critique before labeling the summary (see example critique in Figure \ref{fig:critique}). This type of support can help annotators think of factors they may have missed otherwise but workers may also overrely on the LLM to do the work for them.

\begin{figure}[h]
    \centering
    \includegraphics[width=\columnwidth]{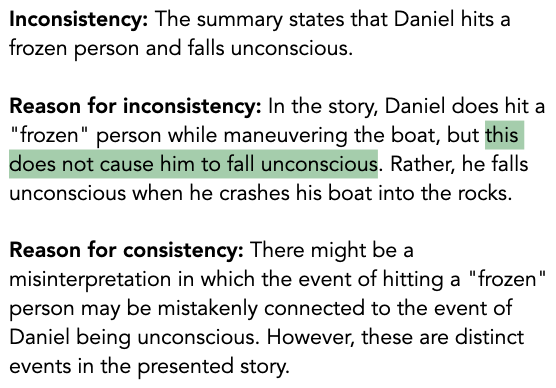}
    \caption{An example of GPT-4's generated inconsistency/support/refute argument for the \textit{debate} method of evaluation. The sentence highlighted in green correctly argues the inconsistency.}
    \label{fig:debate}
\end{figure}

\paragraph*{Debate.} Prior work has shown that having a model present both sides of an issue can help mitigate biasing a human to an LLM decision \cite{si2023large, michael2023debate}. We therefore have GPT-4 generate a possible inconsistency between the summary and story and arguments both for and against this inconsistency. Five workers from Upwork read the inconsistency and these arguments before labeling the summary (see example of inconsistency and arguments in \ref{fig:debate}).

\begin{table*}
\centering \small
\begin{tabular}{l|r|c|ll|rrr}
Method & Coh.-k & \% Faith. & Prec. & Rec. & \% Easy & \% Hard & BAcc.\\ \hline\hline
Binary (Claude-3) & 0.21 & 91 & 0.6 & 1.0 & 11.1 & 33.3 & 60.0 \\
Binary (GPT-4) & 0.27 & 48 & 0.69 & 0.61 & 66.7 & 66.7 & 63.9 \\
Binary (Mixtral) & 0.01 & 94 & 0.55 & 0.94 & 11.1 & 0.0 & 50.6 \\
CoT (Claude-3) & 0.28 & 88 & 0.62 & 1.0 & 22.2 & 33.3 & 63.3 \\
CoT (GPT-4) & 0.22 & 85 & 0.61 & 0.94 & 22.2 & 33.3 & 60.6 \\
CoT (Mixtral) & 0.01 & 94 & 0.55 & 0.94 & 11.1 & 0.0 & 50.6 \\\hline
FABLES (GPT-4) & 0.23 & 39 & 0.69 & 0.5 & 77.8 & 66.7 & 61.7 \\
MiniCheck (Flan-T5) & -0.13 & 12 & 0.25 & 0.06 & 88.9 & 66.7 & 42.8 \\
UniEval & 0.38 & 61 & 0.7 & 0.78 & 66.7 & 50.0 & 68.9 \\
AlignScore & 0.28 & 42 & 0.71 & 0.56 & 77.8 & 66.7 & 64.4 \\\hline
Binary (Human) & 0.44 & 64 & 0.71 & 0.83 & 55.6 & 66.7 & 71.7 \\
Expert & 0.34 & 45 & 0.73 & 0.61 & 88.9 & 50.0 & 67.2 \\
Hybrid & 0.35 & 33 & 0.82 & 0.5 & 88.9 & 83.3 & 68.3 \\
Critique & 0.32 & 64 & 0.67 & 0.78 & 55.6 & 50.0 & 65.6 \\
Debate & 0.04 & 42 & 0.57 & 0.44 & 55.6 & 66.7 & 52.2 
\end{tabular}
\caption{Model scores against the annotator labels on just the validation split.} 
\label{tab:moregoldval}
\end{table*}

\begin{table*}
\centering \small
\begin{tabular}{l|r|c|ll|rrr}
Method & Coh.-k & \% Faith. & Prec. & Rec. & \% Easy & \% Hard & BAcc.\\ \hline\hline
Binary (Claude-3) & 0.13 & 97 & 0.7 & 1.0 & 40.0 & 0.0 & 55.0 \\
Binary (GPT-4) & 0.21 & 83 & 0.73 & 0.88 & 60.0 & 20.0 & 59.2 \\
Binary (Mixtral) & 0.13 & 97 & 0.7 & 1.0 & 0.0 & 13.3 & 55.0 \\
CoT (Claude-3) & 0.19 & 90 & 0.72 & 0.95 & 20.0 & 20.0 & 57.7 \\
CoT (GPT-4) & 0.07 & 98 & 0.69 & 1.0 & 20.0 & 0.0 & 52.5 \\
CoT (Mixtral) & 0.07 & 98 & 0.69 & 1.0 & 0.0 & 6.7 & 52.5 \\\hline
FABLES (GPT-4) & 0.35 & 60 & 0.82 & 0.72 & 60.0 & 66.7 & 68.5 \\
MiniCheck (Flan-T5) & -0.05 & 21 & 0.62 & 0.19 & 80.0 & 73.3 & 46.8 \\
UniEval & -0.09 & 59 & 0.65 & 0.56 & 20.0 & 40.0 & 45.4 \\
AlignScore & -0.0 & 70 & 0.68 & 0.7 & 40.0 & 26.7 & 49.9 \\\hline
Expert & 0.35 & 60 & 0.82 & 0.72 & 100.0 & 53.3 & 68.5 \\
Hybrid & 0.13 & 32 & 0.8 & 0.37 & 100.0 & 73.3 & 58.6 \\
\end{tabular}
\caption{Model scores against the annotator labels on just the test split.} 
\label{tab:moregoldtest}
\end{table*}

\begin{table*}
\centering \small
\begin{tabular}{l|r|c|ll|rrr}
Method & Coh.-k & \% Faith. & Prec. & Rec. & \% Easy & \% Hard & BAcc.\\ \hline\hline
Binary (Claude-3) & 0.06 & 91 & 0.27 & 1.0 & 20.0 & 6.7 & 56.0 \\
Binary (GPT-4) & 0.14 & 48 & 0.31 & 0.62 & 70.0 & 46.7 & 59.2 \\
Binary (Mixtral) & 0.04 & 94 & 0.26 & 1.0 & 10.0 & 6.7 & 54.0 \\
CoT (Claude-3) & 0.08 & 88 & 0.28 & 1.0 & 30.0 & 6.7 & 58.0 \\
CoT (GPT-4) & 0.02 & 85 & 0.25 & 0.88 & 30.0 & 6.7 & 51.7 \\
CoT (Mixtral) & 0.04 & 94 & 0.26 & 1.0 & 10.0 & 6.7 & 54.0 \\\hline
FABLES (GPT-4) & 0.25 & 39 & 0.38 & 0.62 & 70.0 & 66.7 & 65.3 \\
MiniCheck (Flan-T5) & 0.01 & 12 & 0.25 & 0.12 & 80.0 & 93.3 & 50.2 \\
UniEval & 0.13 & 61 & 0.3 & 0.75 & 70.0 & 26.7 & 59.5 \\
AlignScore & 0.21 & 42 & 0.36 & 0.62 & 80.0 & 53.3 & 63.3 \\\hline
Binary (Human) & 0.31 & 64 & 0.38 & 1.0 & 70.0 & 33.3 & 74.0 \\
Expert & 0.55 & 45 & 0.53 & 1.0 & 100.0 & 53.3 & 86.0 \\
Hybrid & 0.63 & 33 & 0.64 & 0.88 & 100.0 & 73.3 & 85.8 \\
Annotator & 0.42 & 55 & 0.44 & 1.0 & 100.0 & 33.3 & 80.0 \\
Critique & -0.01 & 64 & 0.24 & 0.62 & 70.0 & 13.3 & 49.2 \\
Debate & 0.21 & 42 & 0.36 & 0.62 & 60.0 & 66.7 & 63.3 \\
\end{tabular}
\caption{Model scores against the expanded gold labels on just the validation split.} 
\label{tab:moreexpval}
\end{table*}

\begin{table*}
\centering \small
\begin{tabular}{l|r|c|ll|rrr}
Method & Coh.-k & \% Faith. & Prec. & Rec. & \% Easy & \% Hard & BAcc.\\ \hline\hline
Binary (Claude-3) & 0.05 & 97 & 0.46 & 1.0 & 20.0 & 0.0 & 52.9 \\
Binary (GPT-4) & 0.05 & 83 & 0.46 & 0.86 & 40.0 & 12.0 & 52.9 \\
Binary (Mixtral) & 0.05 & 97 & 0.46 & 1.0 & 10.0 & 4.0 & 52.9 \\
CoT (Claude-3) & 0.1 & 90 & 0.47 & 0.96 & 10.0 & 16.0 & 55.4 \\
CoT (GPT-4) & 0.03 & 98 & 0.45 & 1.0 & 10.0 & 0.0 & 51.4 \\
CoT (Mixtral) & 0.03 & 98 & 0.45 & 1.0 & 0.0 & 4.0 & 51.4 \\\hline
FABLES (GPT-4) & 0.26 & 60 & 0.55 & 0.75 & 70.0 & 44.0 & 63.2 \\
MiniCheck (Flan-T5) & -0.12 & 21 & 0.31 & 0.14 & 80.0 & 72.0 & 44.3 \\
UniEval & -0.09 & 59 & 0.41 & 0.54 & 20.0 & 44.0 & 45.4 \\
AlignScore & 0.09 & 70 & 0.48 & 0.75 & 50.0 & 28.0 & 54.6 \\\hline
Expert & 0.69 & 60 & 0.74 & 1.0 & 100.0 & 60.0 & 85.7 \\
Hybrid & 0.34 & 32 & 0.7 & 0.5 & 100.0 & 76.0 & 66.4 \\
Annotator & 0.54 & 68 & 0.65 & 1.0 & 100.0 & 40.0 & 78.6 \\
\end{tabular}
\caption{Model scores against the expanded gold labels on just the test split.} 
\label{tab:moreexptest}
\end{table*}

\end{document}